\pdfoutput=1

\documentclass[11pt]{article}

\usepackage[final]{acl}

\usepackage{times}
\usepackage{latexsym}

\usepackage[T1]{fontenc}

\usepackage[utf8]{inputenc}

\usepackage{microtype}

\usepackage{inconsolata}

\usepackage{graphicx}

\usepackage{booktabs}

\usepackage{amsmath}
\usepackage{amssymb}
\usepackage{tabularx}
\usepackage{bbold}
\usepackage{subcaption}
\usepackage{caption}
\usepackage{comment}
\usepackage{pifont}
\usepackage{wasysym}

\usepackage{color}
\usepackage{colortbl}
\usepackage{multirow}

\usepackage{float}
\usepackage{mdframed}
\usepackage{tcolorbox}
\usepackage{cleveref}
\usepackage{array}
\usepackage{authblk}   

\usepackage[accsupp]{axessibility}  

\usepackage{orcidlink}

\definecolor{weakorange}{RGB}{255,230,200} 
\definecolor{weakgray}{RGB}{240,240,240} 
\definecolor{colbest}{rgb}{0.1, 0.6, 0.1}
\definecolor{colworst}{rgb}{0.75, 0, 0}
\definecolor{oursrow}{HTML}{EBF5F8}

\usepackage{hyperref}

\usepackage{abbr}

\usepackage{soul}

\definecolor{yellowhighlight}{HTML}{FFF2CC}
\definecolor{purplehighlight}{HTML}{efdce5}
\definecolor{lightpinkhighlight}{HTML}{FBE5E5}
\definecolor{lightgreenhighlight}{HTML}{E6FAE0}

\newcommand{\hlyellow}[1]{\sethlcolor{yellowhighlight}\hl{#1}}
\newcommand{\hlpurple}[1]{\sethlcolor{purplehighlight}\hl{#1}}
\newcommand{\hlpink}[1]{\sethlcolor{lightpinkhighlight}\hl{#1}}
\newcommand{\hlgreen}[1]{\sethlcolor{lightgreenhighlight}\hl{#1}}

\title{From Descriptive Richness to Bias: Unveiling the Dark Side of \\ Generative Image Caption Enrichment}

\author{Yusuke Hirota$^{1}$\thanks{Work partially done as an intern at NVIDIA Research.}, Ryo Hachiuma$^{2}$, Chao-Han Huck Yang$^{2}$, Yuta Nakashima$^{1}$ \\
Osaka University$^{1}$\quad NVIDIA Research$^{2}$\\
{\tt\small \{y-hirota@is., n-yuta@\}ids.osaka-u.ac.jp, \{rhachiuma,hucky\}@nvidia.com}\\
}

\begin{document}
\maketitle

\begin{abstract}
Large language models (LLMs) have enhanced the capacity of vision-language models to caption visual text. This generative approach to image caption enrichment further makes textual captions more descriptive, improving alignment with the visual context. However, while many studies focus on benefits of generative caption enrichment (GCE), are there any negative side effects? We compare standard-format captions and recent GCE processes from the perspectives of ``gender bias'' and ``hallucination'', showing that enriched captions suffer from increased gender bias and hallucination. Furthermore, models trained on these enriched captions amplify gender bias by an average of $30.9$\% and increase hallucination by $59.5$\%. This study serves as a caution against the trend of making captions more descriptive.
\end{abstract}

\section{Introduction}
\label{sec:intro}

 \begin{figure*}[t]
   \centering
   \includegraphics[clip, width=0.99\textwidth]{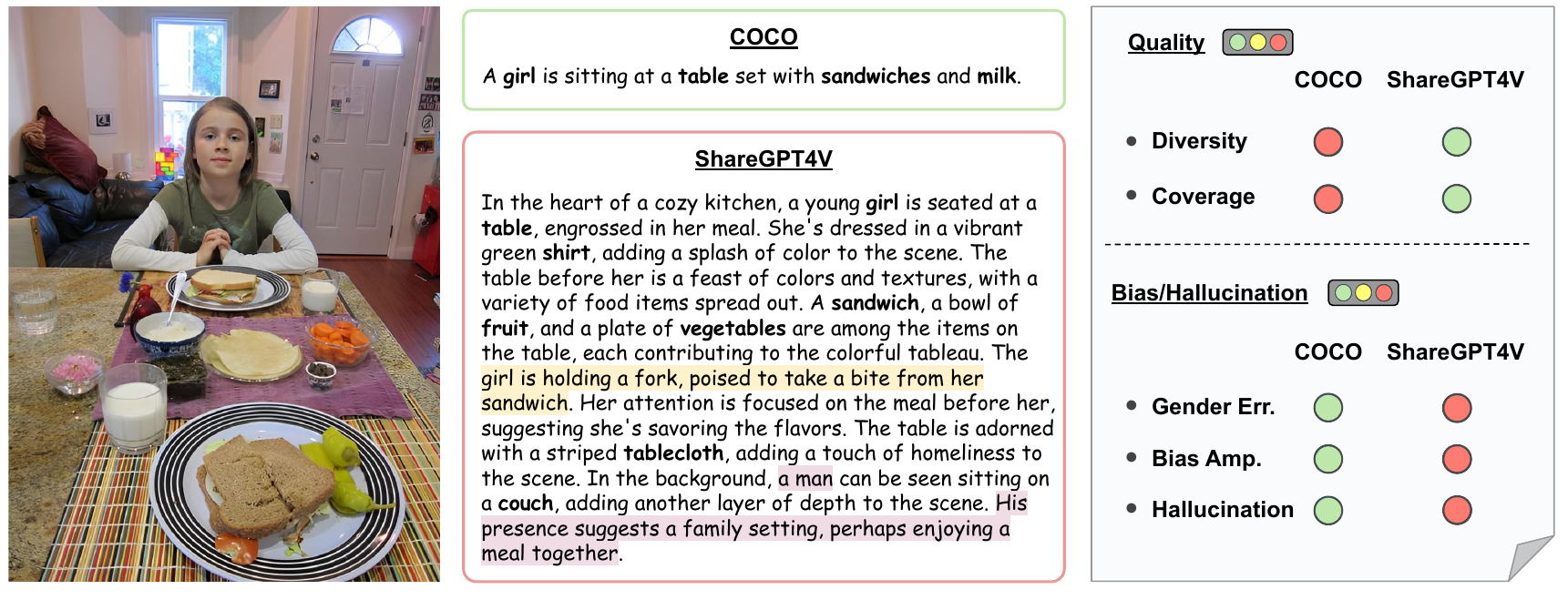}
   \vspace{-5pt}
   \caption{Left: an overview of our analysis. Although the ``LLM-enriched'' caption (ShareGPT4V) covers more content than standard COCO (objects described in captions are \textbf{bolded}), it exhibits hallucination (in \hlyellow{yellow}) and gender bias, including describing gender not exist in the image and possible gender-stereotypical sentence (in \hlpurple{purple}). Right: a comparison between standard and enriched captions on caption quality, bias, and hallucination.}
   \label{fig:fig1}
   \vspace{-5pt}
 \end{figure*}

Large vision-language models (VLMs), such as BLIP \cite{li2023blip}, with superior performance in multi-modal understanding \cite{Luddecke2022CVPR,tewel2022zerocap}, benefiting from millions of image-caption pairs. Improving training paradigms~\cite{wang2024visionllm, liu2024improved} and data augmentation strategies \cite{rotstein2024fusecap, li2024evcap} are crucial topics to enhance VLM performance in image captioning. 

Among these techniques, Generative language models based Caption Enrichment  (GCE) methods \cite{chen2023sharegpt4v, chan2023ic3} have achieved some of the latest state-of-the-art performances. Unlike standard caption benchmarks, which concisely describe the salient parts~\cite{misra2016seeing} of an image (\eg, COCO captions \cite{chen2015microsoft}), GCE methods create more descriptive and semantically enhanced captions. These enhanced textual captions are aligned to boost downstream performance with large language models (LLMs).

While many studies emphasize improving caption quality, issues such as \textit{societal bias} and \textit{hallucination} are significant yet often overlooked~\cite{zhou2023analyzing,zhao2021captionbias,wang2021biasamp,hirota2023model,burns2018women} in image captioning. For example, \citet{zhao2021captionbias} found that the COCO dataset is \textit{skewed} towards men, and \citet{hirota2022quantifying} showed that models trained on this biased data generate gender-stereotypical captions (\eg, describing a \textit{pink dress} for women not wearing one). These studies have highlighted potential biases in datasets like COCO and the models trained on them. Addressing this bias is crucial as it can exacerbate unfairness and risks towards underrepresented groups.

We aim to examine one critical question that has been overlooked in GCE works: \textit{``Although LLM-enriched captions boost VLM performance, do they have negative effects, regarding societal bias and object hallucination?''} To answer this, we investigate gender bias and hallucination using comprehensive metrics, examining both datasets and models trained on these datasets for standard captions (COCO captions) and enriched captions (ShareGPT4V \cite{chen2023sharegpt4v}, FuseCap \cite{rotstein2024fusecap}, CapsFusion \cite{yu2023capsfusion}).

Our analysis reveals that LLM-enriched captions indeed have negative side effects, worsening issues of gender bias and hallucination by making captions more descriptive. Meanwhile, models trained on these enriched captions tend to amplify these problems further. Finally, we discuss possible causes of these negative effects and warn against the trend of making captions more descriptive.

\section{Evaluation Framework}
\label{sec:framework}


\subsection{Generative Caption Enrichment (GCE)}
\label{sec:dataset}

We introduce recent representative approaches to generate enriched captions: ShareGPT4V \cite{chen2023sharegpt4v}, FuseCap \cite{rotstein2024fusecap}, and Capsfusion \cite{yu2023capsfusion}. All these GCE methods utilize LLMs \cite{brown2020language} or Large Multi-modal Models (LMMs) \cite{2023GPT4VisionSC} to describe images in detail or summarize different sources of the information.

\noindent
\textbf{ShareGPT4V} 
utilizes GPT4-Vision \cite{2023GPT4VisionSC} to generate $1.2$M large scale high-quality captions for incremental training on a strong 7B VLM with strong generalization and SOTA results.

\noindent
\textbf{FuseCap}
uses several pre-trained off-the-shelf vision models (\eg, object detector) to extract diverse visual information. The outputs from these models and original captions are fused using ChatGPT \cite{ouyang2022training} to generate enriched captions.

\noindent
\textbf{Capsfusion}
generates captions using a pre-trained captioner, BLIP \cite{li2023blip}, then fuses them with original captions using ChatGPT. 

\begin{table*}[t]
\renewcommand{\arraystretch}{1.1}
\setlength{\tabcolsep}{5pt}
\footnotesize
\caption{Caption quality, gender bias, and hallucination for upstream and downstream analysis. \textcolor{colworst}{Red}/\textcolor{colbest}{green} indicates the worst/best score for each metric. Recall, gender bias, and hallucination metrics are scaled by 100.}
\centering
\vspace{-7pt}
\begin{tabularx}{0.91\textwidth}{l r r r r r r r r r r}
\toprule
& \multicolumn{3}{c}{Caption Quality $\uparrow$} &&\multicolumn{3}{c}{Gender bias $\downarrow$} &&\multicolumn{2}{c}{Hallucination $\downarrow$}\\ 
\cline{2-4} 
\cline{6-8}
\cline{10-11}
Captions & \multirow{1.3}{*}{Diversity} & \multirow{1.3}{*}{Length} & \multirow{1.3}{*}{Recall} && \multirow{1.3}{*}{Gender Err.} & \multirow{1.3}{*}{LIC} & \multirow{1.3}{*}{Recall Disp.} && \multirow{1.3}{*}{CHAIR\textsubscript{s}}  & \multirow{1.3}{*}{CHAIR\textsubscript{i}}  \\
\midrule
\textit{\textbf{Upstream}} &  &  &  &  &  & & & & & \\
COCO captions & \textcolor{colworst}{\textbf{12,834}} & \textcolor{colworst}{\textbf{11.3}} & \textcolor{colworst}{\textbf{42.6}} && \textcolor{colbest}{\textbf{0}} & \textcolor{colbest}{\textbf{0}} & \textcolor{colbest}{\textbf{7.0}} && \textcolor{colbest}{\textbf{0}} & \textcolor{colbest}{\textbf{0}}  \\
ShareGPT4V  & 25,349 & \textcolor{colbest}{\textbf{166.1}} & \textcolor{colbest}{\textbf{61.7}} && 2.5 & \textcolor{colworst}{\textbf{17.4}} & \textcolor{colworst}{\textbf{24.9}} && \textcolor{colworst}{\textbf{20.7}} & \textcolor{colworst}{\textbf{5.7}} \\
FuseCap  & \textcolor{colbest}{\textbf{25,892}} & 39.8 & 59.4 && \textcolor{colworst}{\textbf{3.2}} & 14.3 & 9.9 && 9.2 & 4.0\\
CapsFusion  & 13,158 & 16.9 & 44.6 && 1.4 & 1.2 & 7.6 && 3.5 & 2.2\\
\midrule
\textit{\textbf{Downstream}} &  &  &  &  &  & & & & &\\
COCO captions & \textcolor{colworst}{\textbf{3,312}} & \textcolor{colworst}{\textbf{10.9}} & \textcolor{colworst}{\textbf{45.7}} && \textcolor{colbest}{\textbf{3.1}} & \textcolor{colbest}{\textbf{5.5}} & \textcolor{colbest}{\textbf{7.8}} && \textcolor{colbest}{\textbf{4.7}} & \textcolor{colbest}{\textbf{3.1}}\\
ShareGPT4V  & \textcolor{colbest}{\textbf{9,573}} & \textcolor{colbest}{\textbf{153.8}} & 56.3 && 3.4 & 14.3 & \textcolor{colworst}{\textbf{30.5}} && \textcolor{colworst}{\textbf{21.5}} & \textcolor{colworst}{\textbf{6.9}}\\
FuseCap  & 6,341 & 42.0 & \textcolor{colbest}{\textbf{56.9}} && \textcolor{colworst}{\textbf{4.8}} & \textcolor{colworst}{\textbf{17.3}} & 16.3 && 13.2 & 6.3\\
CapsFusion  & 3,385 & 15.3 & 48.0 && 4.2 & 6.3 & 8.8 && 7.2 & 4.4\\
\bottomrule
\end{tabularx}
\label{tab:main}
\vspace{-5pt}
\end{table*}

\subsection{Evaluation metrics}
\label{sec:metrics}

Our analysis focuses on caption quality, societal bias, and hallucination. Let $\mathcal{D}$ be a dataset of $n$ samples, $\mathcal{D} = \{(I_i, c_i, a_i) \mid 1 \leq i \leq n\}$, where each sample includes an image $I_i$, a caption $c_i$, and an optional binary gender label $a_i$ (\texttt{woman} or \texttt{man}). We introduce the metrics to evaluate each aspect.

\textbf{Caption quality.} 
We evaluate caption quality in three aspects: \textbf{\textit{Vocabulary diversity}} is the total number of unique words across all captions in $\mathcal{D}$. \textbf{\textit{Caption length}} is the average number of tokens per caption in $\mathcal{D}$. \textbf{\textit{Recall}} measures the proportion of objects mentioned in captions to the total objects in the images. For each caption $c_i \in \mathcal{D}$, recall is calculated as:
\begin{equation}
\text{Recall} = \frac{1}{n} \sum_{i=1}^{n} \frac{r_i}{o_i},
\end{equation}
where $o_i$ is the total number of objects in $I_i$, and $r_i$ is the number of relevant objects mentioned in $c_i$. Note that conventional reference-based metrics like CIDEr \cite{vedantam2015cider} cannot be applied to descriptive captions \cite{chan2023ic3}.

\textbf{Societal bias.}
We focus on gender bias as gender terms are more frequently described in captions than other attributes. We adopt three metrics to measure gender bias:
\textbf{\textit{Gender error}} \cite{burns2018women} measures the rate of incorrect gender predictions in captions. If a caption $c_i \in \mathcal{D}$ with gender label $a_i$ refers to a woman as a \textit{man} or vice versa, it counts as an error. The gender error is the proportion of such errors in $\mathcal{D}$. 
\textbf{\textit{Recall disparity}} \cite{hall2023vision}
evaluates the recall disparity between genders. Consider two subsets based on $a_i$: $\mathcal{D}_\text{woman}$ and $\mathcal{D}_\text{man}$. Recall disparity is the average absolute difference in recall for each object $j$:
\begin{equation}
    \resizebox{0.88\linewidth}{!}{
    $\text{Disparity} = \frac{1}{m} \sum_{j=1}^{m} |\text{Recall}_{\text{man},j} - \text{Recall}_{\text{woman},j}|$
    }
\end{equation}
where $m$ is the total number of COCO objects \cite{lin2014microsoft}, $\text{Recall}_{\text{man},j}$ is the recall of COCO objects $j$ in $\mathcal{D}_\text{man}$, and vice versa.
\textbf{\textit{LIC}} \cite{hirota2022quantifying}
quantifies how gender-stereotypical captions in $\mathcal{D}$ are compared to human-written captions. It compares the accuracies of two gender classifiers: one trained on $c_i \in \mathcal{D}$ and the other on ground-truth captions. Higher accuracy for the classifier trained on $c_i$ indicates more gender-stereotypical information in these captions. 

\textbf{Hallucination.}
We use the CHAIR metric \cite{rohrbach2018object} to evaluate hallucination in captions. CHAIR has two components: \textbf{\textit{$\text{CHAIR}_{i}$}} is the fraction of mentioned objects in the captions $c_i$ that do not appear in images $I_i$:
\begin{equation}
    \text{CHAIR}_\text{i} = \frac{H}{M},
\end{equation}
where $H$ is the number of hallucinated objects, and $M$ is the total number of objects mentioned in the captions.
\textbf{\textit{$\text{CHAIR}_{s}$}} is the fraction of captions $c_i$ with at least one hallucinated object:
\begin{equation}
    \text{CHAIR}_\text{s} = \frac{S_h}{n},
\end{equation}
where $S_h$ is the number of captions with hallucinated objects.
We focus on $80$ objects in COCO. 

\section{Evaluation}
\label{sec:exp}

 \begin{figure*}[t]
   \centering
   \includegraphics[clip, width=0.95\textwidth]{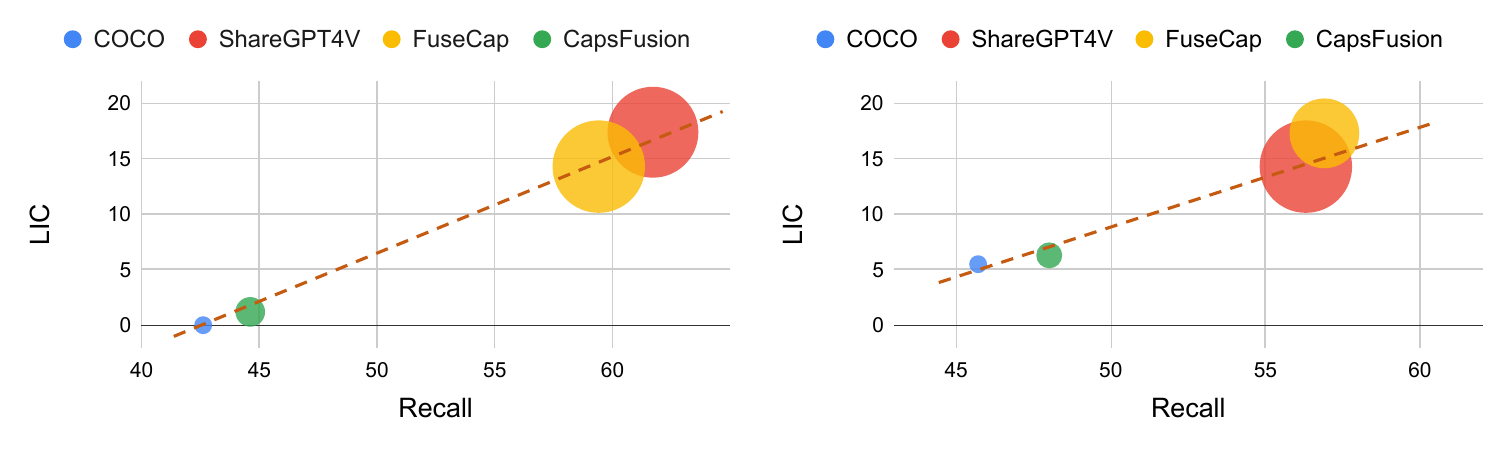}
   \vspace{-15pt}
   \caption{LIC vs. Recall (left: upstream, right: downstream). The bubble size indicates vocabulary size. LIC tends to increase with higher recall, shown by strong trends (dotted lines) with $R^2 = 0.99$ (left) and $R^2 = 0.97$ (right).}
   \label{fig:lic_recall}
   \vspace{-5pt}
 \end{figure*}

  \begin{figure*}[t]
   \centering
   \includegraphics[clip, width=0.95\textwidth]{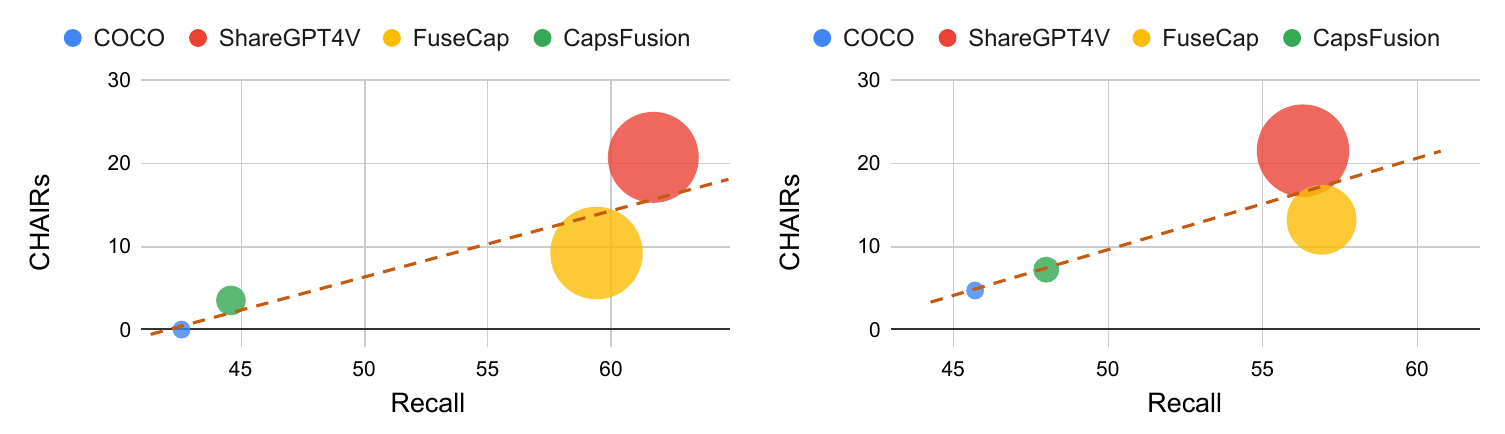}
   \vspace{-11pt}
   \caption{$\text{CHAIR}_\text{s}$ vs. Recall (left: upstream, right: downstream). The bubble size indicates vocabulary size. $\text{CHAIR}_\text{s}$ tends to increase with higher recall, shown by strong trends with $R^2 = 0.80$ (left) and $R^2 = 0.76$ (right).}
   \label{fig:hall_recall}
   \vspace{-5pt}
 \end{figure*}

 \begin{table}[t]
\renewcommand{\arraystretch}{1.1}
\setlength{\tabcolsep}{5pt}
\footnotesize
\caption{Difference in gender bias and hallucination between upstream and downstream (downstream - upstream). \hlpink{Red}/\hlgreen{green} is bias amplification/mitigation.}
\centering
\vspace{-7pt}
\begin{tabularx}{0.99\columnwidth}{l r r r r r r}
\toprule
& \multicolumn{3}{c}{$\Delta$Gender bias } &&\multicolumn{2}{c}{$\Delta$Hallucination}\\ 
\cline{2-4} 
\cline{6-7}
Captions & \multirow{1.3}{*}{Err.} & \multirow{1.3}{*}{LIC} & \multirow{1.3}{*}{Disp.} && \multirow{1.3}{*}{C\textsubscript{s}}  & \multirow{1.3}{*}{C\textsubscript{i}}  \\
\midrule
COCO cap.  & \hlpink{3.1} & \hlpink{5.5} & \hlpink{0.8} && \hlpink{4.7} & \hlpink{3.1} \\
ShareGPT4V  & \hlpink{0.9} & \hlgreen{-3.1} & \hlpink{5.6} && \hlpink{0.8} & \hlpink{1.2} \\
FuseCap & \hlpink{1.6} & \hlpink{3.0} & \hlpink{6.4} && \hlpink{4.0} & \hlpink{2.3} \\
CapsFusion  & \hlpink{2.8} & \hlpink{5.1} & \hlpink{1.2} && \hlpink{3.7} & \hlpink{2.2} \\
\bottomrule
\end{tabularx}
\label{tab:amplification}
\vspace{-5pt}
\end{table}

\paragraph{Setup.}

We analyze concise (COCO captions) and enriched captions (ShareGPT4V, FuseCap, CapsFusion) based on the metrics in \Cref{sec:metrics}. Enriched captions are generated for the COCO training set using these approaches. We first conduct an \textit{upstream} analysis of the four datasets and then a \textit{downstream} analysis of captions generated by a captioner trained on each dataset. For downstream analysis, we fine-tune a pre-trained BLIP for 5 epochs with the AdamW optimizer, generating captions for the COCO validation set. Detailed experimental settings are in \Cref{app:exp}.

\subsection{Upstream \& downstream analysis}
\label{sec:analysis}

  \begin{figure}[t]
   \centering
   \includegraphics[clip, width=0.9\columnwidth]{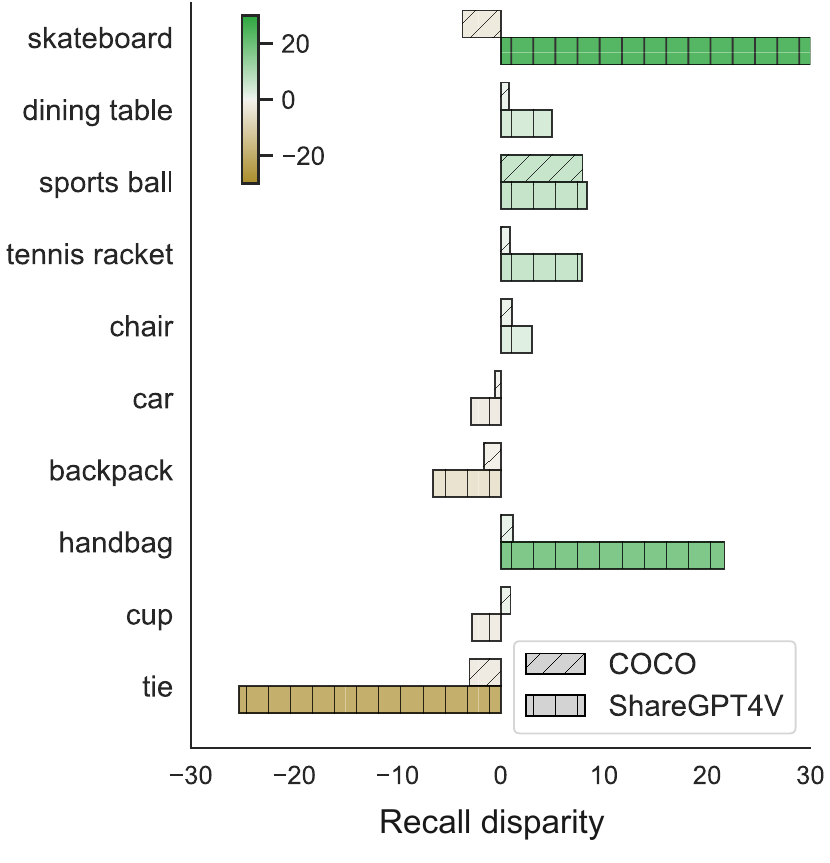}
   \vspace{-8pt}
   \caption{Recall disparity by visual object.}
   \label{fig:recall_obj}
   \vspace{-5pt}
 \end{figure}

 We present qualitative results in \Cref{fig:fig1} and \Cref{app:qualitative}, with key observations  below.

\textbf{\textit{Observation 1.} More descriptive, more gender bias.}
\Cref{tab:main} (upstream) shows a clear tendency for gender bias to increase as captions become more descriptive. For instance, COCO captions have the lowest object coverage (\ie, recall: 42.6) but exhibit the least bias. In contrast, ShareGPT4V and FuseCap have higher object coverage but higher gender bias than COCO captions (\eg, LIC is 0 for COCO and 17.4 for ShareGPT4V). This observation is further confirmed by \Cref{fig:lic_recall} (left), showing a strong correlation between LIC and recall ($R^2 = $ 0.99). In other words, making captions more descriptive increases the risk of gender bias.

\textbf{\textit{Observation 2.} Enriched captions exhibit greater recall disparity.}
In \Cref{fig:recall_obj}, we visualize the difference in recall ($\text{Recall}_{\text{man}} - \text{Recall}_{\text{woman}}$) for the top-$10$ objects that co-occur with images in $\mathcal{D}_\text{woman}$ and $\mathcal{D}_\text{man}$. The results show that ShareGPT4V exhibits a more significant recall disparity for all objects. For example, for the \textit{handbag} object, COCO captions show almost no gender difference, while ShareGPT4V exhibits a strong bias towards men. This further validates the risk of gender bias in enriched captions.

\textbf{\textit{Observation 3.} More descriptive, more hallucination.}
A similar trend between descriptiveness and hallucination is also evident in \Cref{tab:main} (upstream). COCO captions, which has the lowest object coverage, exhibits the lowest hallucination rates. Conversely, ShareGPT4V, with the highest object coverage, shows significantly increased hallucination rates compared to COCO captions (\eg, $\text{CHAIR}_{s}$ is 0 for COCO and 20.7 for ShareGPT4V). This trend is corroborated by \Cref{fig:hall_recall} (left), highlighting a strong correlation between hallucination rates and recall ($R^2 = $ 0.80). Thus, making captions more descriptive increases hallucination risks.

\textbf{\textit{Observation 4.} Models trained on the datasets inherit/amplify bias and hallucination.}
\Cref{tab:main} (downstream) shows that models inherit the dataset's bias tendencies. Specifically, the model trained on the least descriptive captions (\ie, COCO captions) exhibits the smallest bias and hallucination, while models trained on the most descriptive captions, ShareGPT4V and FuseCap, show significant bias and hallucination. \Cref{fig:lic_recall,fig:hall_recall} (right) further demonstrate that the models inherit the datasets' bias and hallucination. Furthermore, \Cref{tab:amplification} shows that in most cases, the models amplify the dataset's biases. For example, ShareGPT4V's recall disparity worsens from $24.9$ to $30.5$ ($\Delta = 5.6$), and $\text{CHAIR}_{s}$ increases from $20.7$ to $21.5$ ($\Delta = 0.8$). These results highlight the severe issue of dataset bias, as it directly affects the outcomes of the trained models.

 \section{Discussion on Possible Sources of Bias}
 \label{sec:discussion}
 To enhance descriptiveness, GEC methods heavily rely on LLMs to improve textual alignment. However, issues with gender bias and hallucination \cite{gunjal2024detecting} have been explored in these LLMs. The enrichment process, which depends on text representations, risks incorporating these inherent biases into the final captions. Furthermore, the lack of human oversight in the caption generation process can exacerbate these issues. Without iterative human intervention to correct biases, the inaccuracies of LLMs remain unaddressed, leading to increased bias and hallucination. Introducing human-in-the-loop~\cite{yang2019study} could mitigate these problems by ensuring that captions are free from gender-stereotypical descriptions.

\section{Conclusion}
\label{sec:conclusion}
We examined standard and LLM-enriched captions for gender bias and hallucination, deriving key insights: GCE-based image captioning exacerbates these bias, which are further amplified in downstream models. We argue that further efforts must be invested to the problems to strike a balance between \textit{descriptive richness} and \textit{incremental bias}.

\section*{Limitations}
\label{sec:limitations}

\paragraph{Attributes other than gender.}
We focused our analysis on gender bias for societal bias. This is because gender-related terms are more frequently described in captions compared to other attributes, making gender bias particularly prominent in captions \cite{hirota2022quantifying}. However, previous works have shown that racial bias, though not as pronounced as gender bias, is also present in captioning models \cite{zhao2021captionbias}. Analyzing racial bias and bias of other attributes requires future studies and efforts.

\paragraph{Evaluation metrics.}
While our analysis demonstrated various critical problems in enriched captions (\eg, they exacerbate bias and hallucination), there may be aspects that we can further investigate. For example, we can consider other attributes for societal bias analysis and utilize hallucination metrics that account for elements \textit{beyond objects}. However, our analysis is robust and highlights critical issues in enriched captions, serving as a counterpoint to the trend of making captions more descriptive and benefiting the community.

\paragraph{Source datasets other than COCO.}
In our analysis, we used COCO as the source for images for two reasons: (1) COCO images come with high-quality, human-annotated concise captions, providing a solid basis for evaluating concise captions; and (2) COCO has been extensively analyzed in existing research for societal bias and hallucination \cite{li2023halueval}. We did not use other image-caption datasets (\eg, Google Conceptual Captions \cite{sharma2018conceptual}, LAION \cite{schuhmann2022laion}) because the quality of the accompanying captions is lower, making the analysis results less reliable.

\section*{Acknowledgments}

We thank David Chan from UC Berkeley and Yen-Ting Lin from National Taiwan University for their invaluable contributions during the initial discussions that  enhanced this project.

\bibliography{custom}

\begin{thebibliography}{34}
\providecommand{\natexlab}[1]{#1}

\bibitem[{Brown et~al.(2020)Brown, Mann, Ryder, Subbiah, Kaplan, Dhariwal, Neelakantan, Shyam, Sastry, Askell et~al.}]{brown2020language}
Tom Brown, Benjamin Mann, Nick Ryder, Melanie Subbiah, Jared~D Kaplan, Prafulla Dhariwal, Arvind Neelakantan, Pranav Shyam, Girish Sastry, Amanda Askell, et~al. 2020.
\newblock Language models are few-shot learners.
\newblock In \emph{NeurIPS}.

\bibitem[{Burns et~al.(2018)Burns, Hendricks, Saenko, Darrell, and Rohrbach}]{burns2018women}
Kaylee Burns, Lisa~Anne Hendricks, Kate Saenko, Trevor Darrell, and Anna Rohrbach. 2018.
\newblock Women also snowboard: Overcoming bias in captioning models.
\newblock In \emph{ECCV}.

\bibitem[{Chan et~al.(2023)Chan, Myers, Vijayanarasimhan, Ross, and Canny}]{chan2023ic3}
David Chan, Austin Myers, Sudheendra Vijayanarasimhan, David~A Ross, and John Canny. 2023.
\newblock Ic3: Image captioning by committee consensus.
\newblock In \emph{EMNLP}.

\bibitem[{Chen et~al.(2023)Chen, Li, Dong, Zhang, He, Wang, Zhao, and Lin}]{chen2023sharegpt4v}
Lin Chen, Jisong Li, Xiaoyi Dong, Pan Zhang, Conghui He, Jiaqi Wang, Feng Zhao, and Dahua Lin. 2023.
\newblock Sharegpt4v: Improving large multi-modal models with better captions.
\newblock \emph{arXiv preprint arXiv:2311.12793}.

\bibitem[{Chen et~al.(2015)Chen, Fang, Lin, Vedantam, Gupta, Doll{\'a}r, and Zitnick}]{chen2015microsoft}
Xinlei Chen, Hao Fang, Tsung-Yi Lin, Ramakrishna Vedantam, Saurabh Gupta, Piotr Doll{\'a}r, and C~Lawrence Zitnick. 2015.
\newblock Microsoft coco captions: Data collection and evaluation server.
\newblock \emph{arXiv preprint arXiv:1504.00325}.

\bibitem[{Devlin et~al.(2015)Devlin, Cheng, Fang, Gupta, Deng, He, Zweig, and Mitchell}]{devlin2015language}
Jacob Devlin, Hao Cheng, Hao Fang, Saurabh Gupta, Li~Deng, Xiaodong He, Geoffrey Zweig, and Margaret Mitchell. 2015.
\newblock Language models for image captioning: The quirks and what works.
\newblock In \emph{Proceedings of the 53rd Annual Meeting of the Association for Computational Linguistics and the 7th International Joint Conference on Natural Language Processing (Volume 2: Short Papers)}. Association for Computational Linguistics.

\bibitem[{Gunjal et~al.(2024)Gunjal, Yin, and Bas}]{gunjal2024detecting}
Anisha Gunjal, Jihan Yin, and Erhan Bas. 2024.
\newblock Detecting and preventing hallucinations in large vision language models.
\newblock In \emph{AAAI}.

\bibitem[{Hall et~al.(2023)Hall, Gustafson, Adcock, Misra, and Ross}]{hall2023vision}
Melissa Hall, Laura Gustafson, Aaron Adcock, Ishan Misra, and Candace Ross. 2023.
\newblock Vision-language models performing zero-shot tasks exhibit gender-based disparities.
\newblock In \emph{ICCV Workshops}.

\bibitem[{Hirota et~al.(2022)Hirota, Nakashima, and Garcia}]{hirota2022quantifying}
Yusuke Hirota, Yuta Nakashima, and Noa Garcia. 2022.
\newblock Quantifying societal bias amplification in image captioning.
\newblock In \emph{CVPR}.

\bibitem[{Hirota et~al.(2023)Hirota, Nakashima, and Garcia}]{hirota2023model}
Yusuke Hirota, Yuta Nakashima, and Noa Garcia. 2023.
\newblock Model-agnostic gender debiased image captioning.
\newblock In \emph{CVPR}.

\bibitem[{Li et~al.(2024)Li, Vo, Sugimoto, and Nakayama}]{li2024evcap}
Jiaxuan Li, Duc~Minh Vo, Akihiro Sugimoto, and Hideki Nakayama. 2024.
\newblock Evcap: Retrieval-augmented image captioning with external visual-name memory for open-world comprehension.
\newblock In \emph{CVPR}.

\bibitem[{Li et~al.(2023{\natexlab{a}})Li, Li, Savarese, and Hoi}]{li2023blip}
Junnan Li, Dongxu Li, Silvio Savarese, and Steven Hoi. 2023{\natexlab{a}}.
\newblock Blip-2: Bootstrapping language-image pre-training with frozen image encoders and large language models.
\newblock \emph{arXiv preprint arXiv:2301.12597}.

\bibitem[{Li et~al.(2023{\natexlab{b}})Li, Cheng, Zhao, Nie, and Wen}]{li2023halueval}
Junyi Li, Xiaoxue Cheng, Wayne~Xin Zhao, Jian-Yun Nie, and Ji-Rong Wen. 2023{\natexlab{b}}.
\newblock Halueval: A large-scale hallucination evaluation benchmark for large language models.
\newblock In \emph{EMNLP}.

\bibitem[{Lin et~al.(2014)Lin, Maire, Belongie, Hays, Perona, Ramanan, Doll{\'a}r, and Zitnick}]{lin2014microsoft}
Tsung-Yi Lin, Michael Maire, Serge Belongie, James Hays, Pietro Perona, Deva Ramanan, Piotr Doll{\'a}r, and C~Lawrence Zitnick. 2014.
\newblock Microsoft {COCO}: Common objects in context.
\newblock In \emph{ECCV}.

\bibitem[{Liu et~al.(2024)Liu, Li, Li, and Lee}]{liu2024improved}
Haotian Liu, Chunyuan Li, Yuheng Li, and Yong~Jae Lee. 2024.
\newblock Improved baselines with visual instruction tuning.
\newblock In \emph{CVPR}.

\bibitem[{Louizos et~al.(2017)Louizos, Shalit, Mooij, Sontag, Zemel, and Welling}]{louizos2017causal}
Christos Louizos, Uri Shalit, Joris~M Mooij, David Sontag, Richard Zemel, and Max Welling. 2017.
\newblock Causal effect inference with deep latent-variable models.
\newblock \emph{Advances in neural information processing systems}, 30.

\bibitem[{L\"uddecke and Ecker(2022)}]{Luddecke2022CVPR}
Timo L\"uddecke and Alexander Ecker. 2022.
\newblock Image segmentation using text and image prompts.
\newblock In \emph{CVPR}.

\bibitem[{Misra et~al.(2016)Misra, Lawrence~Zitnick, Mitchell, and Girshick}]{misra2016seeing}
Ishan Misra, C~Lawrence~Zitnick, Margaret Mitchell, and Ross Girshick. 2016.
\newblock Seeing through the human reporting bias: Visual classifiers from noisy human-centric labels.
\newblock In \emph{CVPR}.

\bibitem[{OpenAI(2023)}]{2023GPT4VisionSC}
OpenAI. 2023.
\newblock \href {https://api.semanticscholar.org/CorpusID:263218031} {Gpt-4v(ision) system card}.
\newblock In \emph{OpenAI Blog}.

\bibitem[{Ouyang et~al.(2022)Ouyang, Wu, Jiang, Almeida, Wainwright, Mishkin, Zhang, Agarwal, Slama, Ray et~al.}]{ouyang2022training}
Long Ouyang, Jeffrey Wu, Xu~Jiang, Diogo Almeida, Carroll Wainwright, Pamela Mishkin, Chong Zhang, Sandhini Agarwal, Katarina Slama, Alex Ray, et~al. 2022.
\newblock Training language models to follow instructions with human feedback.
\newblock In \emph{NeurIPS}.

\bibitem[{Rohrbach et~al.(2018)Rohrbach, Hendricks, Burns, Darrell, and Saenko}]{rohrbach2018object}
Anna Rohrbach, Lisa~Anne Hendricks, Kaylee Burns, Trevor Darrell, and Kate Saenko. 2018.
\newblock Object hallucination in image captioning.
\newblock In \emph{EMNLP}.

\bibitem[{Rotstein et~al.(2024)Rotstein, Bensa{\"\i}d, Brody, Ganz, and Kimmel}]{rotstein2024fusecap}
Noam Rotstein, David Bensa{\"\i}d, Shaked Brody, Roy Ganz, and Ron Kimmel. 2024.
\newblock Fusecap: Leveraging large language models for enriched fused image captions.
\newblock In \emph{WACV}.

\bibitem[{Schuhmann et~al.(2022)Schuhmann, Beaumont, Vencu, Gordon, Wightman, Cherti, Coombes, Katta, Mullis, Wortsman et~al.}]{schuhmann2022laion}
Christoph Schuhmann, Romain Beaumont, Richard Vencu, Cade Gordon, Ross Wightman, Mehdi Cherti, Theo Coombes, Aarush Katta, Clayton Mullis, Mitchell Wortsman, et~al. 2022.
\newblock Laion-5b: An open large-scale dataset for training next generation image-text models.
\newblock In \emph{NeurIPS}.

\bibitem[{Sharma et~al.(2018)Sharma, Ding, Goodman, and Soricut}]{sharma2018conceptual}
Piyush Sharma, Nan Ding, Sebastian Goodman, and Radu Soricut. 2018.
\newblock Conceptual captions: A cleaned, hypernymed, image alt-text dataset for automatic image captioning.
\newblock In \emph{ACL}.

\bibitem[{Tewel et~al.(2022)Tewel, Shalev, Schwartz, and Wolf}]{tewel2022zerocap}
Yoad Tewel, Yoav Shalev, Idan Schwartz, and Lior Wolf. 2022.
\newblock Zerocap: Zero-shot image-to-text generation for visual-semantic arithmetic.
\newblock In \emph{CVPR}.

\bibitem[{Vedantam et~al.(2015)Vedantam, Lawrence~Zitnick, and Parikh}]{vedantam2015cider}
Ramakrishna Vedantam, C~Lawrence~Zitnick, and Devi Parikh. 2015.
\newblock Cider: Consensus-based image description evaluation.
\newblock In \emph{CVPR}.

\bibitem[{Vinyals et~al.(2015)Vinyals, Toshev, Bengio, and Erhan}]{vinyals2015show}
Oriol Vinyals, Alexander Toshev, Samy Bengio, and Dumitru Erhan. 2015.
\newblock Show and tell: A neural image caption generator.
\newblock In \emph{CVPR}.

\bibitem[{Wang and Russakovsky(2021)}]{wang2021biasamp}
Angelina Wang and Olga Russakovsky. 2021.
\newblock Directional bias amplification.
\newblock In \emph{ICML}.

\bibitem[{Wang et~al.(2023)Wang, Chen, Chen, Wu, Zhu, Zeng, Luo, Lu, Zhou, Qiao et~al.}]{wang2024visionllm}
Wenhai Wang, Zhe Chen, Xiaokang Chen, Jiannan Wu, Xizhou Zhu, Gang Zeng, Ping Luo, Tong Lu, Jie Zhou, Yu~Qiao, et~al. 2023.
\newblock Visionllm: Large language model is also an open-ended decoder for vision-centric tasks.
\newblock In \emph{NeurIPS}.

\bibitem[{Yang et~al.(2022)Yang, Hung, Ouyang, and Chen}]{yang2022training}
Chao-Han~Huck Yang, I-Te~Danny Hung, Yi~Ouyang, and Pin-Yu Chen. 2022.
\newblock Training a resilient q-network against observational interference.
\newblock In \emph{AAAI}.

\bibitem[{Yang et~al.(2019)Yang, Kandogan, Li, Sen, and Lasecki}]{yang2019study}
Yiwei Yang, Eser Kandogan, Yunyao Li, Prithviraj Sen, and Walter~S Lasecki. 2019.
\newblock A study on interaction in human-in-the-loop machine learning for text analytics.
\newblock In \emph{IUI Workshops}.

\bibitem[{Yu et~al.(2024)Yu, Sun, Zhang, Cui, Zhang, Wang, and Liu}]{yu2023capsfusion}
Qiying Yu, Quan Sun, Xiaosong Zhang, Yufeng Cui, Fan Zhang, Xinlong Wang, and Jingjing Liu. 2024.
\newblock Capsfusion: Rethinking image-text data at scale.
\newblock In \emph{CVPR}.

\bibitem[{Zhao et~al.(2021)Zhao, Wang, and Russakovsky}]{zhao2021captionbias}
Dora Zhao, Angelina Wang, and Olga Russakovsky. 2021.
\newblock Understanding and evaluating racial biases in image captioning.
\newblock In \emph{ICCV}.

\bibitem[{Zhou et~al.(2023)Zhou, Cui, Yoon, Zhang, Deng, Finn, Bansal, and Yao}]{zhou2023analyzing}
Yiyang Zhou, Chenhang Cui, Jaehong Yoon, Linjun Zhang, Zhun Deng, Chelsea Finn, Mohit Bansal, and Huaxiu Yao. 2023.
\newblock Analyzing and mitigating object hallucination in large vision-language models.
\newblock In \emph{ICLR}.

\end{thebibliography}

\appendix

\clearpage
\onecolumn
\section{Appendix Settings of the Experiments}
\label{app:exp}
\subsection{Upstream analysis}
For the upstream analysis, we use images of the COCO training set as an image source. For every image, we apply GCE approaches introduced in \Cref{sec:dataset}, obtaining enriched captions for each approach. We compare these enriched captions with concise captions (COCO captions) in terms of the metrics introduced in \Cref{sec:metrics}.

\subsection{Downstream analysis}
For the downstream analysis, we fine-tune pre-trained BLIP on one of the four datasets (\ie, COCO captions, ShareGPT4V, FuseCap, or CapsFusion). We fine-tune the models for 5 epochs with the AdamW optimizer. Then, the fine-tuned models are used to generate captions for the COCO validation set. The generated captions for the validation set are used for the downstream analysis. \

 \begin{figure*}[ht!]
   \centering
   \includegraphics[clip, width=0.95\textwidth]{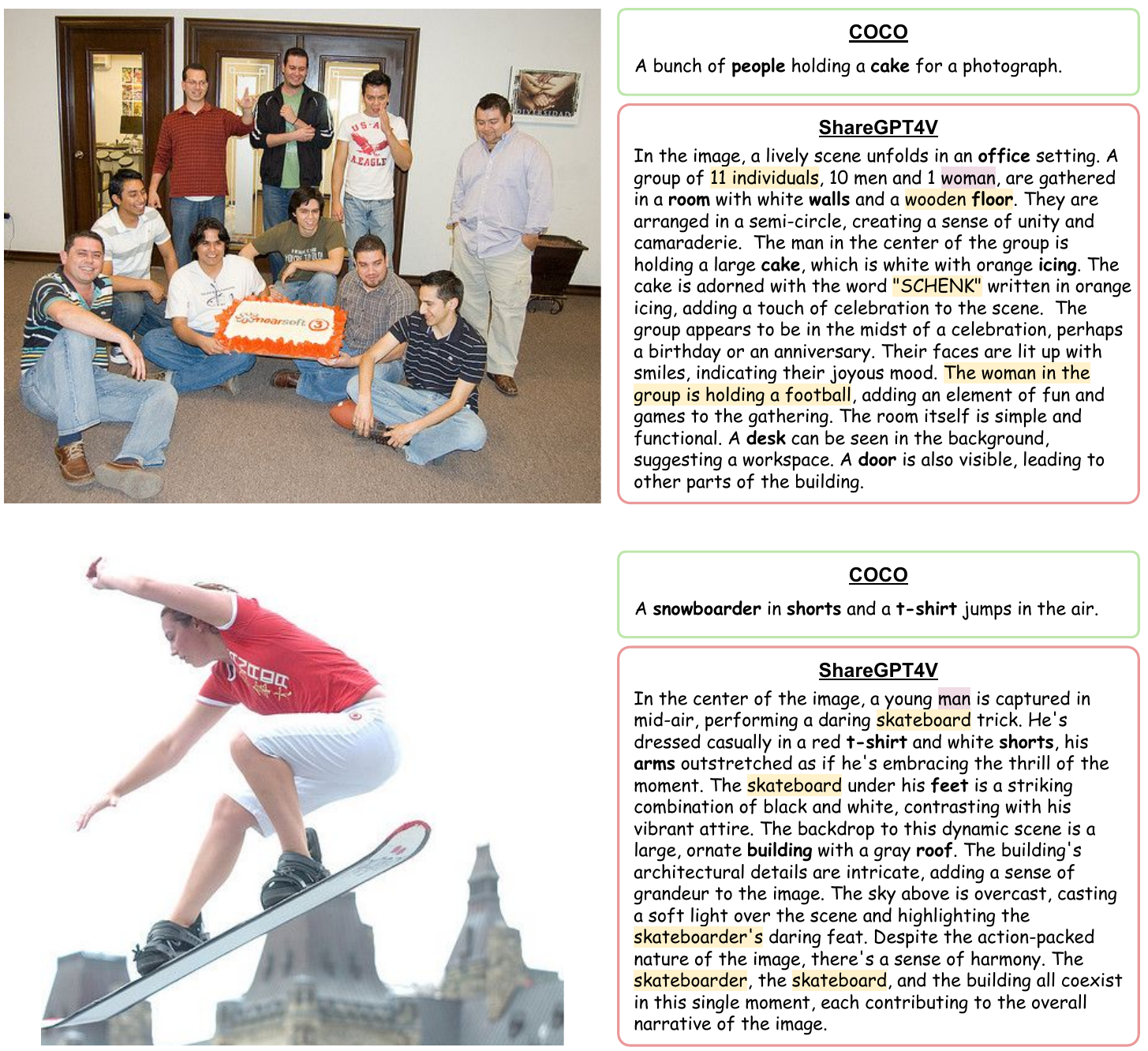}
   \vspace{-3pt}
   \caption{Qualitative examples of the comparison between COCO captions and ShareGPT4V. Objects described in captions are \textbf{bolded}. Gender bias and hallucination are highlighted in \hlpurple{purple} and \hlyellow{yellow}, respectively.}
   \label{fig:sharegpt-qualitative}
 \end{figure*}

  \begin{figure*}[ht!]
   \centering
   \includegraphics[clip, width=0.95\textwidth]{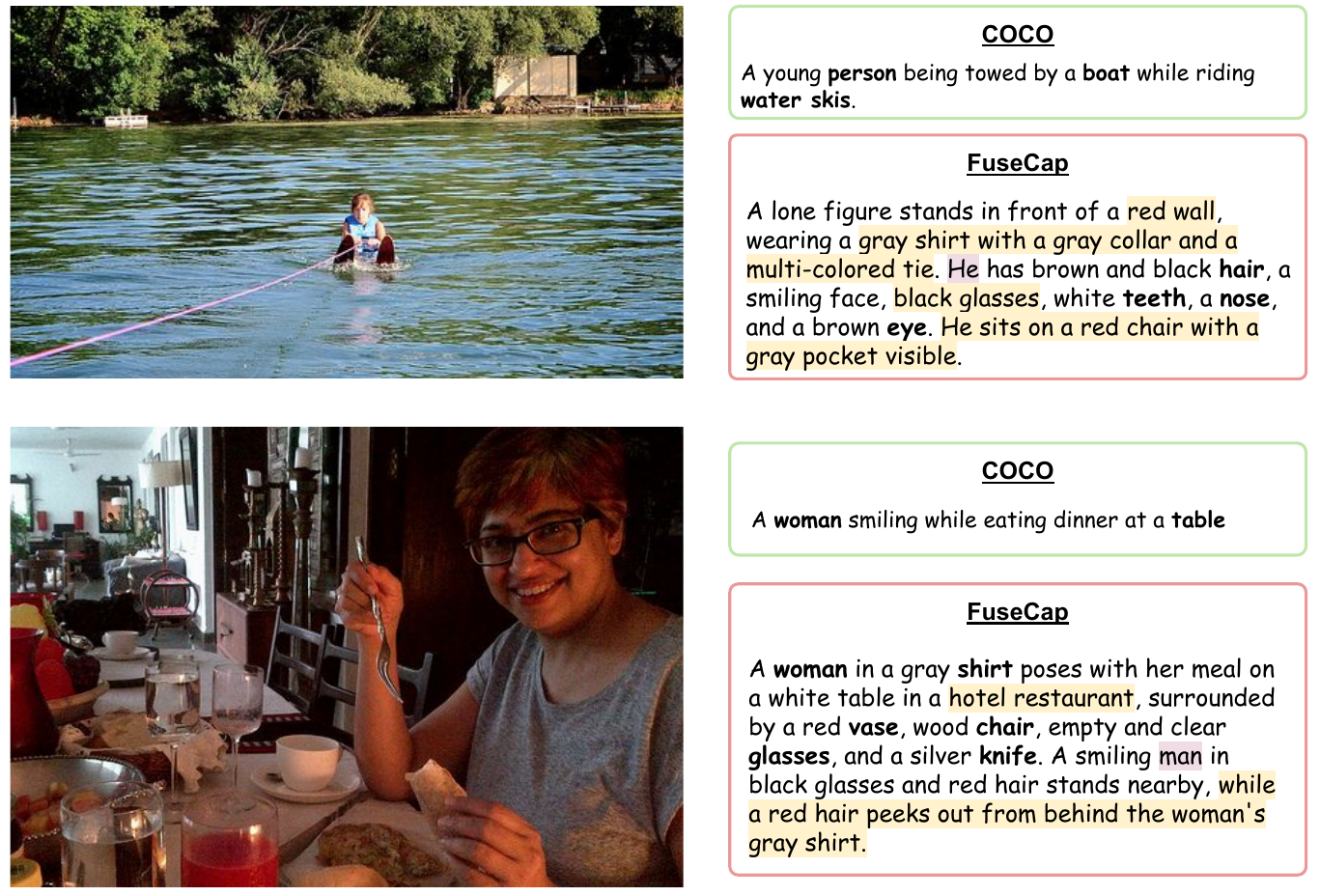}
   \vspace{-3pt}
   \caption{Qualitative examples of the comparison between COCO captions and FuseCap. Objects described in captions are \textbf{bolded}. Gender bias and hallucination are highlighted in \hlpurple{purple} and \hlyellow{yellow}, respectively.}
   \label{fig:fusecap-qualitative}
 \end{figure*}

 \begin{figure*}[ht!]
   \centering
   \includegraphics[clip, width=0.95\textwidth]{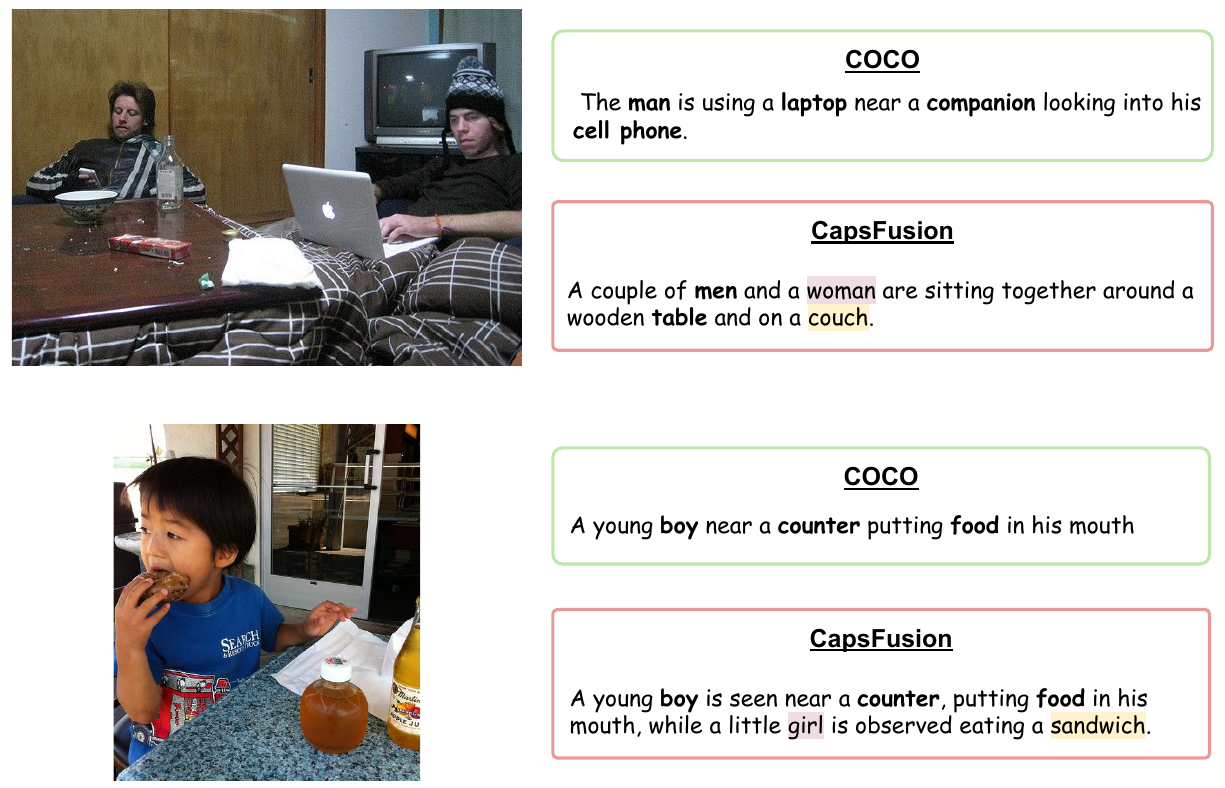}
   \vspace{-3pt}
   \caption{Qualitative examples of the comparison between COCO captions and CapsFusion. Objects described in captions are \textbf{bolded}. Gender bias and hallucination are highlighted in \hlpurple{purple} and \hlyellow{yellow}, respectively. }
   \label{fig:capsfusion-qualitative}
 \end{figure*}

\section{Additional Qualitative Examples}
\label{app:qualitative}

 We show some qualitative examples where enriched captions generated by GCE methods exhibit gender bias and hallucination. The examples are shown in \Cref{fig:sharegpt-qualitative,fig:fusecap-qualitative,fig:capsfusion-qualitative} (ShareGPT4V, FuseCap, and CapsFusion, respectively). 

The enriched captions suffer from gender misclassification (\eg, bottom of \Cref{fig:sharegpt-qualitative}), incorrectly describing two people in the image as a couple (\eg, top of \Cref{fig:capsfusion-qualitative}), describing nonexistent individuals with different genders (\eg, bottom of \Cref{fig:fusecap-qualitative}), and object hallucination (in all the figures). These results further confirm the negative impacts of GCE.

\subsection{Non-LLM based Caption Enrichment}
\label{app:non-llm}

We also would like to credit previous works~\cite{devlin2015language, vinyals2015show} of non pre-training based language modeling to enhance image captioning by providing structured linguistic patterns and vocabulary. However, without the depth of large language models, such systems may exhibit bias and limited expressiveness, struggling to generate diverse and contextually nuanced captions. These models often rely on statistical techniques, which can constrain their descriptive capabilities compared to their more advanced counterparts. In other words, how to incorporate structure knowledge refinement~\cite{louizos2017causal,yang2022training} or graphical structure would also be important for LLM-based caption enrichment in future studies.

\end{document}